# Ensemble Classifiers and Their Applications: A Review

Akhlaqur Rahman[1] and Sumaira Tasnim[2]
[1]Department of Electrical and Electronic Engineering Uttara University, Bangladesh
[2]School of Engineering, Deakin University, Australia

*ABSTRACT:*

*Ensemble classifier refers to a group of individual classifiers that are cooperatively trained on data set in a supervised classification problem. In this paper we present a review of commonly used ensemble classifiers in the literature. Some ensemble classifiers are also developed targeting specific applications. We also present some application driven ensemble classifiers in this paper.*

## 1. INTRODUCTION

A supervised classification problem falls under the category of learning from instances where each instance/pattern/example is associated with a label/class. Conventionally an individual classifier like Neural Network, Decision Tree, or a Support Vector Machine is trained on a labeled data set. Depending on the distribution of the patterns, it is possible that not all the patterns are learned well by an individual classifier. Classifier performs poorly on the test set under such scenarios.

A solution to this problem is to train a group of classifiers (Fig. 1) on the same problem. The existing literature coined the term 'Ensemble Classifier' to refer to the group [1]. Individual classifiers are called base/weak classifiers. During learning, the base classifiers are trained separately on the data set. During prediction, the base classifiers provide a decision on a test pattern. A fusion method then combines the decisions produced by the base classifiers. There exists a good number of fusion methods in the literature including majority voting, Borda count, algebraic combiners etc. [1].

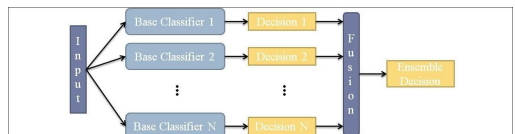

Fig 1: Ensemble of classifiers

The philosophy of the ensemble classifier is that another base classifier compensates the errors made by one base classifier. However, training the base classifier in a straightforward manner is not going to solve this problem. As pointed out in [1] an ensemble classifier performs better than its base counterpart if the base classifiers are accurate and diverse. The term diversity refers to the fact that the base classifier errors be uncorrelated. There are a good number of ways to compute diversity including Pairwise diversity measures (the Q statistics, the correlation coefficient, the disagreement measure, the double fault measure) and non-pairwise diversity measure (the entropy measure, Kohavi-Wolpert varience, measurement of interrater agreement) [2] -[4].

Different ensemble classifier generation methods aim to achieve diversity among the base classifiers. Some ensemble classifiers are also developed targeting specific problems/applications. The following section details different base classifiers, ensemble classifiers, and their applications.

## 2. BASE CLASSIFIERS

Base classifiers refer to individual classifiers used to construct the ensemble classifiers. Neural network, support vector machine, and k-NN classifiers are some of the commonly used base classifiers. For the sake of completeness we briefly explain the training and test process of these base classifiers. In k-NN classification the distance between a test pattern and all the patterns in the training set is computed. The distance can be calculated using Euclidian distance or Manhattan distance. The probable classes receive a vote from each of the k patterns that are closest to the test pattern in terms of distance. The class that obtains the highest vote is considered to be the class of the test pattern.

A neural network [5] can be considered as a computing system made up of a number of simple, highly interconnected processing elements, which process information by their dynamic state response to external inputs. Neural networks are organized in layers. Layers are made up of a number of interconnected nodes that contain an activation function. Patterns are presented to the network via the input layer, that communicates to one or more hidden layers where the actual processing is done via a system of weighted connections. The hidden layers then link to an output layer where the answer is output. Most Neural Networks contain a learning rule which modifies the weights of the connections according to the input patterns that it is presented with.

An SVM [6] classifies data by transforming the data into higher dimension using a kernel function and then finding the best hyperplane that separates the patterns of one class from those of the other class. The best hyperplane for an SVM refers to the one with the maximum margin between the classes. Margin means the maximal width of the slab parallel to the hyperplane that has no interior patterns. The support vectors are the data points that are closest to the separating hyperplane; these points are on the boundary of the slab. In this paper we have used SVM as the base classifier.

### 3. ENSEMBLE CLASSIFIERS

Ensemble classifier generation methods using can be broadly classified into six groups [29] that are based on (i) manipulation of the training parameters, (ii) manipulation of the error function, (iii) manipulation of the feature space, (iv) manipulation of the output labels, (v) clustering, and (vi) manipulation of the training patterns.

3.1 Ensemble Classifier Generation by Manipulation of the Training Parameters

Diversity can be achieved by manipulating the training parameters of the base classifiers in an ensemble. Different network weights are used to train the base neural network learning process in [7] and [8]. These methods achieve better generalization.

3.2 Ensemble Classifier Generation by Manipulation of the Error Function

A group of ensemble classifier construction methods address this issue by augmenting the error function of the base classifiers. An error is imposed if base classifiers make identical errors on similar patterns. Negative correlation learning [9] and [10] is one such ensemble where all the individual networks in the ensemble are trained simultaneously and interactively through the correlation penalty terms in their error functions.

3.3 Ensemble Classifier Generation by Manipulation of The Feature Space

In another group of ensemble classifiers diversity among the base classifiers is achieved by manipulating the input feature space. Different feature subsets are used to train the base classifiers [11][12] and [13]. The random subspace ensemble classifiers perform relatively inferior to other ensemble classifiers.

3.4 Ensemble Classifier Generation by Manipulation of the Output Labels

Ensemble classifiers can be constructed by manipulation of the output targets [14] and [15]. In class switching ensemble [14], each base classifier is generated by switching the class labels of a fraction of training patterns that are selected at random from the original training set.

3.5 Ensemble Classifier Generation by Clustering

Ensemble classifiers can be generated by partitioning the training set into non-overlapping clusters and training base classifiers on them. These classifiers are called clustered ensembles [16]-[20]. The patterns that tend to stay close in Euclidean space naturally are identified by this process. A pattern can belong to one cluster only thus a selection approach is followed for obtaining the ensemble class decision. These methods aim to reduce the learning complexity of large data sets [16]. The clustered ensembles [17]-[20] do not provide any mechanism for obtaining the optimal number of clusters.

Some researchers [21]-[31] provide a mechanism to obtain soft partitioning of the data set that leads to better classification performance [21]. In [25]-[31] ensemble classifier is generated by (i) partitioning data into clusters at different layers, (ii) training base classifiers at clusters at different layers. During prediction (i) the nearest cluster at each layer was found for the test pattern, (ii) prediction was obtained from the classifiers in the nearest clusters at each layer, and (iii) the decisions from the layers are fused into a single decision using majority voting. In [25]-[27] same numbers of clusters are used at different layers and in [31] different numbers of clusters were used. The optimality of the number of clusters and layers are dealt with in [28]-[31].

3.6 Ensemble Classifier Generation by Manipulation of the Training Patterns

The largest set of ensembles generates ensemble classifiers by manipulating the training patterns where the base classifiers are trained on different subsets of the training patterns. The methods differ in generation of the subsets.

In bagging [32] the training subsets are randomly drawn (with replacement) from the training set. Homogeneous base classifiers are trained on the subsets. The class chosen by most base classifiers is the considered to be the final verdict of the ensemble classifier. There are a number of variants of bagging and aggregation approaches including random forests [33] and large scale bagging [34]. Boosting [35] creates data subsets for base classifier training by re-sampling the training patterns, however, by providing the most informative training pattern for each consecutive classifier. Each of the training patterns is assigned a weight that determines how well the instance was classified in the previous iteration. The training data that are wrongly classified is included in the training subset for the next iteration. AdaBoost [36] is a more generalized version of boosting.

## 4. ENSEMBLE CLASSIFIER APPLICATIONS

Ensemble classifiers are sometimes developed targeting specific applications. This section presents some applications of ensemble classifiers.

### 4.1 Sensor Data Quality Assessment

A novel machine learning approach to assess the quality of sensor data using an ensemble classification framework is presented in [37][38]. The quality of sensor data is indicated by discrete quality flags that indicate the level of uncertainty associated with a sensor reading. The nature of sensor data poses some challenges to the classification task. Data of dubious quality exists in such data sets with very small frequency leading to the class imbalance problem. The authors in [37][38] adopt a cluster oriented under-sampling approach. To improve the overall classification accuracy, the approach produces multiple under–sampled training sets using cluster oriented sampling and train base classifiers on each of them. Decisions produced by the base classifiers are fused into a single decision using majority voting. The ensemble classification framework was evaluated by assessing the quality of marine sensor data obtained from sensors situated at Sullivans Cove, Hobart, Australia. Experimental results reveal that the framework agrees with expert judgement with high accuracy and achieves superior classification performance than other state–of–the–art approaches.

### 4.2 Shellfish Farm Closure Prediction and Cause Identification

Shellfish farms must be closed if there is suspected contamination during production to avoid serious health hazards. The authorities monitor a number of environmental and water quality variables through a set of sensors to check the health of shellfish farms and to decide on the closure of the farms. The research presented in [39][40] develops an ensemble of class-balancing classifiers (similar to [37][38]) to identify the cause of closure.

### 4.3 Handwriting Recognition

In [41][42] the authors present novel ensemble classifier architectures and investigate their influence for offline cursive character recognition. Cursive characters are represented by feature sets that portray different aspects of character images for recognition purposes. The recognition accuracy can be improved by training ensemble of classifiers on the feature sets. Given the feature sets and the base classifiers, the authors have developed multiple ensemble classifier compositions under four architectures. The first three architectures are based on the use of multiple feature sets whereas the fourth architecture is based on the use of a unique feature set. Type-1 architecture is composed of homogeneous base classifiers and Type-2 architecture is constructed using heterogeneous base classifiers. Type-3 architecture is based on hierarchical fusion of decisions. In Type-4 architecture a unique feature set is learned by a set of homogeneous base classifiers with different learning parameters. The experimental results demonstrate that the recognition accuracy achieved using Type-4 ensemble classifier is better than the other recognition accuracies for offline cursive character recognition.

### 4.4 Benthic Habitat Mapping

In [43] the authors present a novel approach to produce benthic habitat maps from sea floor images. The authors have developed a step–by–step segmentation method to separate sea–grass, sand, and rock from the sea floor image. The sea–grass was separated first using color filtering. The remaining image was classified into rock and sand based on color, texture, and edge features. The features were fed into an ensemble classifier to produce better classification results. The base classifiers in the ensemble were made complementary by changing the weight (i.e. cost of misclassification) of the classes. The habitat maps were produced for three regions in Derwent estuary. Experimental results demonstrate that the method can indentify different objects and produce habitat maps from the sea–floor images with very high accuracy.

### 4.5 Dealing with Missing Sensor Data

Because of the uncertainty associated with the data acquisition process, a full set of sensor values is not always available for decision making purposes. The prediction system thus needs to deal with missing values. Statistical approaches are commonly used to generate an artificial value to approximate a missing sensor reading and predictions are made on the then complete set of sensor values. In [44][45] the authors present a new method that is capable of making predictions without making artificial approximations of missing values. The idea is to train a set of classifiers on different subsets of sensor values. Given a set of available sensor values, a prediction is made by the classifier trained on the corresponding set of sensor values. The authors have evaluated the system on the data obtained from a number of shellfish farms in Tasmania. Experimental results demonstrate that the proposed method to deal with missing values can predict closures with high accuracy. In this paper the authors assume equal weight for all sensors that may not always hold [46].

### 4.6 Algae Growth Prediction

In [47], the authors present an approach for predicting algae growth through the selection of influential environmental variables. Chlorophyll-a is considered to be an indicator for algal biomass and the authors predict this as a proxy for algae growth. Environmental variables like water temperature, salinity, etc. have influence upon algae growth. Depending on the geographic location, the influence of these environmental variables will vary. Given a set of relevant environmental variables feature selection was performed using a number of algorithms to identify the variables relevant to the growth. An influence matrix-based approach is developed to fuse the decisions from multiple ranking algorithms and select the relevant features. The selected features are then used for predicting algae growth using different regression algorithms to identify their relative strength. The approach is tested on the algae data of Derwent estuary in Tasmania. The experimental results demonstrate that the accuracy of algae growth prediction with influence matrix-based feature selection is superior to using all the features.

### 5. CONCLUSIONS

In this paper we have presented a set of ensemble classifier generation methods. We have also presented some interesting applications of ensemble classifiers. In future we aim to undertake a similar survey on time series ensemble classifiers.